\pdfoutput=1

\documentclass[11pt]{article}

\usepackage[preprint]{acl}

\usepackage{times}
\usepackage{latexsym}

\usepackage[T1]{fontenc}

\usepackage[utf8]{inputenc}

\usepackage{microtype}

\usepackage{inconsolata}

\usepackage{graphicx}
\usepackage{amsmath}
\usepackage{amsfonts}
\usepackage{comment}
\usepackage[export]{adjustbox}
\usepackage{tabularray}
\usepackage{array}
\newcolumntype{P}[1]{>{\centering\arraybackslash}p{#1}}

\usepackage[symbol]{footmisc}

\title{FLoRA: Fused forward-backward adapters for parameter efficient fine-tuning and reducing inference-time latencies of LLMs}

\author{Dhananjaya Gowda$^*$ \quad Seoha Song$^*$ \quad Junhyun Lee \quad Harshith Goka\\
Samsung Research}

\begin{document}
\maketitle
\begin{abstract}
As the large language models (LLMs) grow in size each day, efficient training and fine-tuning has never been as important as nowadays.
This resulted in the great interest in parameter efficient fine-tuning (PEFT), and effective methods including low-rank adapters (LoRA) has emerged. 
Although the various PEFT methods have been studied extensively in the recent years, the greater part of the subject remains unexplored with the huge degree of freedom. 
In this paper, we propose FLoRA, a family of fused forward-backward adapters (FFBA) for parameter-efficient fine-tuning of LLMs on downstream tasks.
The FFBA combine ideas from the popular LoRA and parallel adapters to improve the overall fine-tuning accuracies.
At the same time, latencies are minimized by fusing the forward and backward adapters into existing projection layers of the base model.
Experimental results show that the proposed FFB adapters perform significantly better than the popularly used LoRA in both accuracy and latency for a similar parameter budget.
\end{abstract}
\footnotetext{$^*$ Equal contribution.}

\section{Introduction}
\label{sec:intro}
Large language models (LLMs) have revolutionized the world of artificial intelligence with downstream applications in almost all possible domains that humans can imagine~\cite{openai2023gpt4}.
With an ever-growing demand to accommodate more domains and newer tasks, and with prohibitively high retraining or full fine-tuning (FFT) costs, parameter-efficient fine-tuning (PEFT)~\cite{mangrulkar2022peft} of LLMs using adapters has become the most commonly used approach to add more capabilities to existing LLMs~\cite{houlsby2019peft}.
Also, the performance of LLMs fine-tuned on individual tasks with separate adapters is often better than that of a generalized LLM that can handle multiple tasks.

LLM adapters can be broadly classified into prompt or prefix fine-tuning, serial adapters, and parallel adapters~\cite{hu-etal-2023-llm}.
Among the different types of adapters proposed in the literature, low-rank adapters (LoRA) and their variants are widely used to fine-tune LLMs~\cite{hu2022lora, Li_2021, hu-etal-2023-llm}.
Serial adapters have been one of the earliest adapters used in the literature of neural networks in several domains, including natural language processing~\cite{rebuffi2017serialada, houlsby2019peft}.
However, one main disadvantage of the serial adapters is the sequential nature of these adapter computations, which cannot be easily parallelized along with the base model computations, leading to significant latency overheads as compared to using only the base models.
Low-rank adapters (LoRA), which have become more popular in recent times, are a type of parallel adapters that are attached in parallel to any linear projection rather than at block or higher levels.
The popularity of LoRA and it's variants stem from their simplicity and ability to merge them back into the base model easily.
Prompt or prefix fine-tuning is an alternative way to adapt LLMs to new tasks with minimal compute overhead, but are generally seen to underperform serial or parallel adapters~\cite{hu-etal-2023-llm}.

LoRA adapters provide an efficient way to fine-tune an LLM to new domains/tasks by freezing the base model and updating only a small set of adapter parameters using a small domain or task-specific dataset~\cite{Mao_2024}.
These adapters can be attached to any linear projection layer and can be easily merged into the base model after fine-tuning but before deployment, there by saving on the additional compute that may introduce non-trivial latency during inference.
However, merging these one or more domain-specific adapters into the base model can deteriorate the performance of the LLM on tasks that it was already good at.
One option is to merge and de-merge these adapters into the base model, however, this can introduce significant overheads.
In view of this, it has become a common practice to deploy a common base LLM model with different domain/task specific adapters that can be switched based on needs.
Invoking these adapter computations as separate calls to the GPU kernel operations introduces significant additional latencies (20-50\%) disproportionate to the insignificant additional computations (1-5\%) required in terms of FLOPs.

One way of countering this is to fuse these adapter parameters to the base model and deploy them as one single operation instead of two.
In this paper, we propose a new family of fused forward-backward adapters that build on the advantages of the existing pool of adapters that provides improved latencies as well as better accuracies on down-stream tasks for a given parameter budget.
The main contributions of this paper are as follows:
\begin{itemize}
    \item A new family of fused adapters referred to as fused forward-backward adapters (FFBA) or fused low-rank adapters (FLoRA) is proposed that help reduce inference time latencies when using LLM adapters.
    \item The proposed fused adapters reduce the time per output token (TPOT) overhead of LoRA adapters by 21-30\% for 1B models and 31-48\% for 3B models.
    \item The proposed fused adapters perform significantly better than LoRA on summary and dialogue tasks, while they are on par or marginally better than LoRA on commonsense and math reasoning tasks.
\end{itemize}

\section{Related Work}

It is now a common practice to prepare a foundation LLM and fine-tune it for downstream tasks. 
FFT became more and more challenging as the foundation model grew in size and parameters, and various PEFT methods have been investigated.

Low-rank adapters (LoRA), currently the most commonly used PEFT method, was first introduced in \citet{hu2022lora} based on the hypothesis that weight updates during a downstream task fine-tuning have a low "intrinsic rank." 
It freezes the original weight $W$ and only updates the low-rank weight difference $\Delta W$, where the low-rank is ensured with the decomposition of $\Delta W = AB$. (to be explained in detail in Section~\ref{sec:fusedada})
The low-rank value, $r$, is the key hyperparameter of LoRA.
With the great success of LoRA, many derivative works which improve on various aspects of the LoRA have been published. 
A comprehensive summary of LoRA and its variants is provided in the survey paper, \citet{Mao_2024}.

Here, we introduce an inexhaustive list of LoRA variants. 
A set of works modify the training scheme, for example, using different learning rates for $A$ and $B$ matrices~\cite{hayou2024lora+}, adding residual connections during training and merge during inference~\cite{shi2024reslora}, or freezing the $A$ matrix and training only $B$ matrix to reduce the memory footprint of training~\cite{zhang2023lora}.
There are another group of studies which concentrate on the low-rank value optimization, such as dynamical rank allocation utilizing SVD of updates~\cite{zhang2023adalora}, adaptive parameter addition~\cite{zhang2023increlora}, and using gating techniques during training based on importance and only keep the most important ranks in the end~\cite{ding2023sparse}. 
\citet{meng2025pissa} optimizes the initialization of LoRA matrices, using principal components of the original weight matrix to initialize $A$ and $B$ and use the residual weight as the frozen weight.

While these works aim to optimize the LoRA's performance, they all preserve the basic structure of LoRA. 
We instead investigate on modifying the structure of LoRA itself. 
This is because our main motivation is to suggest an efficient adapter which can maximize the parallelization of GPUs.

Parallel adapters~\cite{hetowards} are modules connected to either or both the attention or feed-forawrd network (FFN) blocks.
As the name suggests, parallel adapters are linked in parallel in the graph, that is, the input is shared with the attention (FFN) block and the output is added to that of the attention (FFN). 
Typically the adapter consists of a feed-forward down projection, nonlinearity, and a feed-forward up projection. 
\citet{hu-etal-2023-llm} thoroughly investigates the parallel adapter and concludes that in optimal settings its performance matches with LoRA of similar parameter budget.

In this paper, we do no rely on a single type of adapter. 
Rather, we build upon the parallel adapters' expressive power and use it to complement LoRA.
First, we modify LoRA with the intension of efficient inference and less latency, with the possibility of performance drop.
Then we minimally apply the parallel adapter to counterbalance the loss in performance.
Details of the overall strategy will follow in the next section. 

PEFT includes other methods such as prefix or prompt-tuning~\cite{li2021prefix, lester2021power, liu2022p}, where task-dependent learnable embeddings are appended at the beginning of the context. 
Series adapters~\cite{houlsby2019parameter, pfeiffer2020mad} serially insert additional trainable modules to the `attention$-$FFN' sequence in a layer. 
Survey papers~\cite{xu2023parameter, balne2024parameter} are available for comprehensive list of PEFT methods. 

\begin{figure}[t]
    \centering    
    \includegraphics[width=\linewidth]{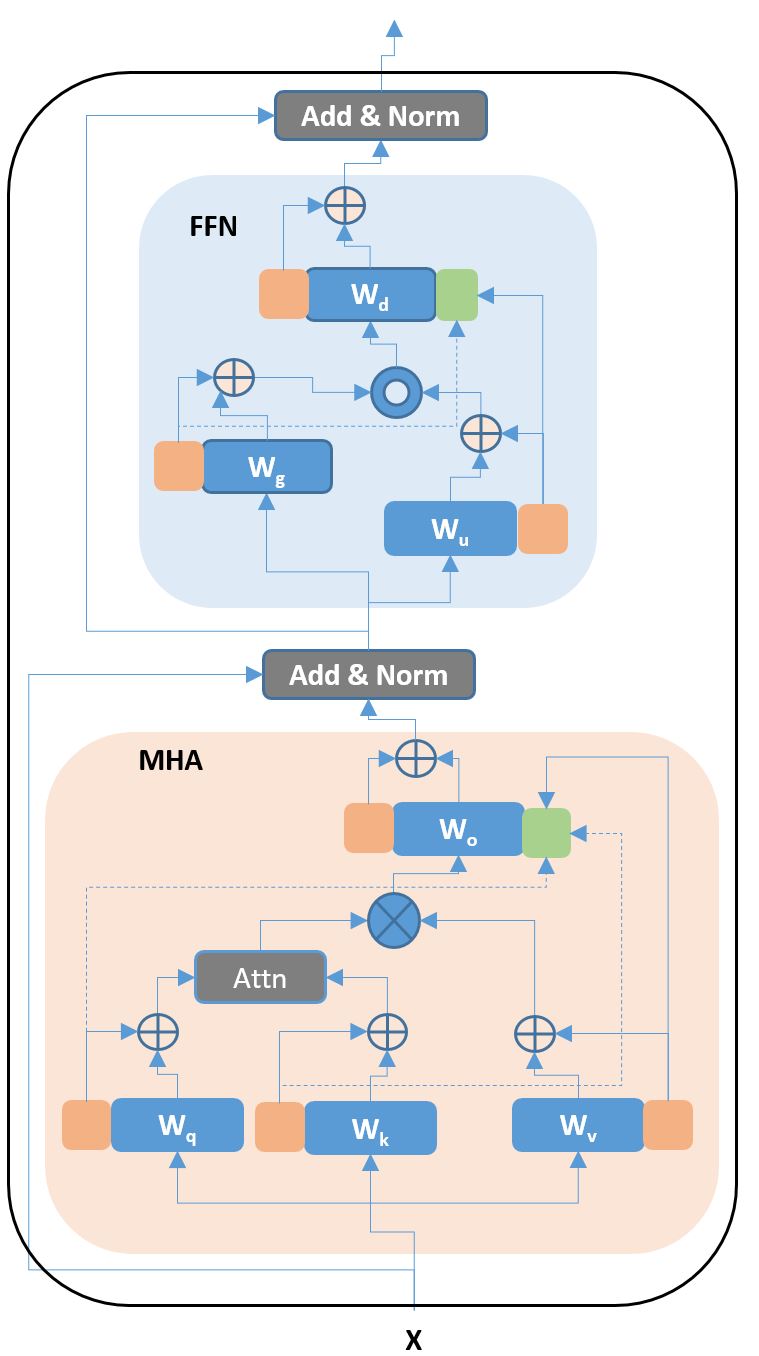}        
    \caption{Block schematic of the proposed fused forward-backward adapter. The blue blocks denote the base model, orange blocks denote the forward adapters, and the green blocks denote the backward adapter.}
    \label{fig:ffba}
\end{figure}
\begin{figure*}[t]
    \centering
    \begin{tabular}{cccc}
    (a) & \includegraphics[width=0.4\linewidth, valign=c]{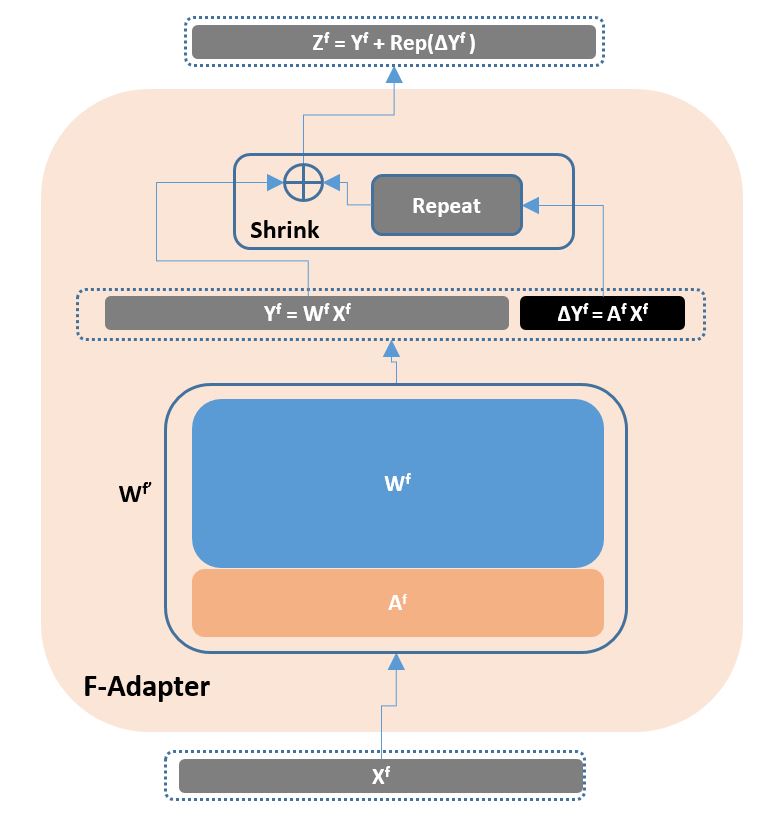} & $\qquad$ (b) &
    \includegraphics[width=0.4\linewidth, valign=c]{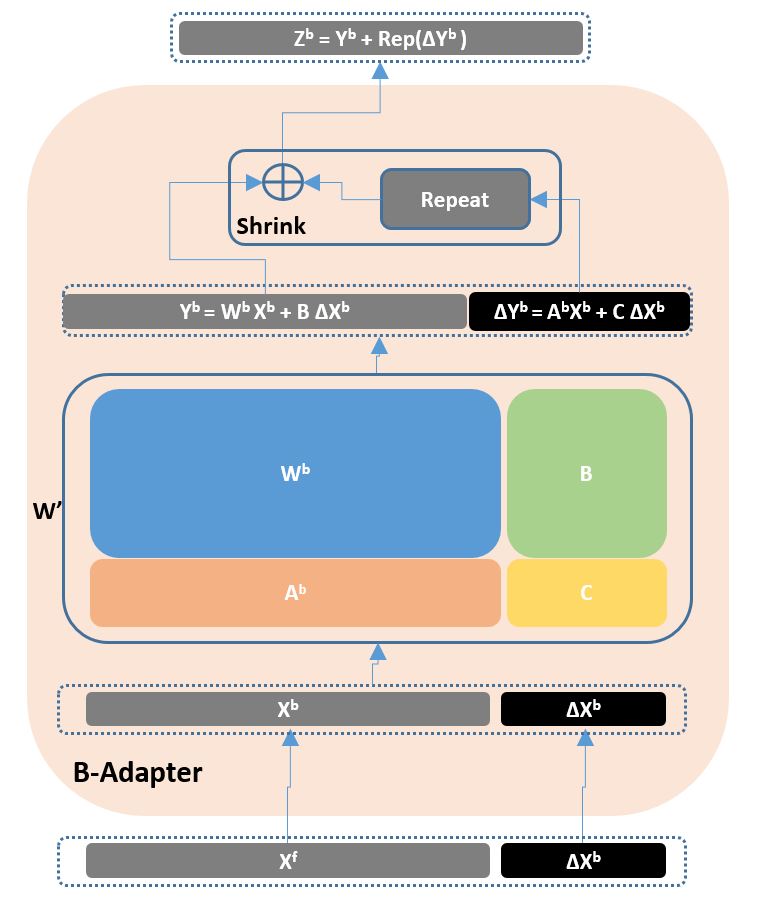}
    \end{tabular}    
    \caption{Linear projection layers with fused forward and backward adapters. (a) Fused forward layer (FFL), and (b) Fused forward-backward layer (FFBL).}
    \label{fig:ffba_blocks}
\end{figure*}
\section{Fused LLM Adapters}
\label{sec:fusedada}
Conventional LoRA uses low-rank approximation (LRA) in order to process information efficiently in a typically large hidden input dimension.
For instance, the output of a linear projection layer with weights $W \in {\mathbb R}^{d_o \times d_i}$ and LoRA adapters $A\in {\mathbb R}^{r \times d_i}$, $B\in {\mathbb R}^{d_o \times r}$, for an input $X\in {\mathbb R}^{d_i \times L}$ is given by
\begin{align}
    Z &= WX + BAX
\end{align}
where $d_i$ and $d_o$ are the input and output dimensions, $L$ is the input sequence length, and $r (\ll d_i$ and $d_o$) is the rank of the LRA of the adapter weight matrix $\Delta W = BA$.
From now on, we refer to the $A$ and $B$ matrices as forward and backward adapters, respectively.

\paragraph{Partially-fused LoRA}
In a naive implementation of LoRA, including the Huggingface PEFT library~\cite{mangrulkar2022peft}, 
the above computation of a single LoRA is performed as a sequence of 4 different operations, namely, $WX$, $AX$, $B(AX)$, and $WX + BAX$.
It is often seen that the overall latency incurred in executing these sequences of operations separately is much larger compared to the total FLOPs that need to be computed.
In order to reduce the overall latency of this compute, and utilize the efficiency of GPUs in parallelization of large size matrix multiplications, the first two operations can be fused into one by concatenating the weight matrcis W and A into one.
The resulting computations are given by
\begin{align}
\begin{bmatrix} Y \\ \Delta Y \end{bmatrix}
= \begin{bmatrix} W \\ A \end{bmatrix} X
= \begin{bmatrix} WX \\ AX \end{bmatrix}
\label{eq:pflora}
\end{align}
where $Y=WX$ and $\Delta Y = AX$.
However, the other two operations $\Delta Z = B\Delta Y$ and $Z = Y+\Delta Z$ still need to computed sequentially.
We refer this way of implementing LoRA may be be referred to as partially-fused LoRA (pf-LoRA).

\paragraph{Fused forward adapter}
One way of further reducing the overall latency is to eliminate the LRA framework and remove the backward projection, $B$.
The saved parameter count can be added to the forward projection matrix $A$ by increasing the low-rank dimension from $r$ to $2 r$.
This may be referred to as fused forward adapter (FFA).
In this case, after calculating Eq.~\ref{eq:pflora} we would need one additional computation $Z = Y + Repeat(\Delta Y)$ in order to combine the concatenated outputs obtained from base $Y$ and adapter $\Delta Y$ weight matrices.
The specific operation used in the combination is a design choice.
In this manuscript, we use ``repeat and add,'' that is, we repeat the $\Delta Y$ vector $d_o/2r$ times to match the dimensions of the two vectors and add them. 

This reduces the overall latency, however, without the LRA bottleneck the ability of the adapter module to effectively capture the additional information may reduce significantly during fine-tuning, as can be seen from experimental results later.

\begin{table*}[h]
    \centering
    {
    \begin{tabular}{l|cccccccc|c}
    \hline
     & \multicolumn{8}{c|}{Commonsense Reasoning Tasks (Acc \%) } \\
         Adapter &  arcc & arce & boolq & hella & obqa & piqa & siqa & wino & Avg\\\hline
         \multicolumn{10}{c}{Llama3.2-1B-Inst} \\\hline
         Base & 51.00 & 73.00 & 64.00 & 44.00 & 74.50 & 72.50 & 50.00 & 45.00 & 59.25 \\         
         FFT  & 59.60 & 75.40 & 81.70 & 73.40 & 84.60 & 78.50 & 73.00 & 71.70 & \bf 74.73 \\\hline
         LoRA & 58.50 & 74.00 & 81.50 & 71.50 & 85.50 & 75.00 & 72.00 & 73.50 & 73.93 \\
         FFA  & 52.50 & 71.00 & 81.50 & 69.50 & 85.00 & 69.50 & 69.50 & 69.50 & 71.00 \\
         FFBA (AorB) & 57.00 & 77.00 & 81.00 & 72.00 & 83.00 & 72.50 & 73.50 & 68.50 & 73.06 \\
         FFBA-Relu (AorB) & 56.00 & 76.50 & 81.50 & 72.50 & 84.00 & 72.00 & 73.50 & 70.50 & 73.31 \\
         FFBA (QG-Add) & 62.10 & 76.00 & 79.90 & 73.40 & 84.60 & 77.70 & 71.70 & 68.90 & \bf 74.28 \\\hline\hline
         \multicolumn{10}{c}{Llama3.2-3B-Inst} \\\hline
         Base & 79.00 & 83.00 & 83.00 & 68.00 & 83.00 & 72.50 & 68.50 & 54.00 & 73.87 \\
         FFT  & 74.50 & 85.00 & 91.50 & 78.50 & 93.50 & 81.00 & 79.50 & 84.00 & \bf 85.68 \\\hline
         LoRA & 78.00 & 86.00 & 90.00 & 84.00 & 93.00 & 86.50 & 84.00 & 84.00 & 84.93 \\
         FFA  & 76.00 & 84.50 & 85.00 & 78.00 & 88.50 & 76.00 & 78.50 & 77.50 & 80.50 \\
         FFBA (AorB) & 81.00 & 89.50 & 89.00 & 85.00 & 94.00 & 78.50 & 78.00 & 82.00 & \bf 85.50 \\
         FFBA-Relu (AorB) & 80.00 & 88.50 & 91.00 & 85.50 & 94.00 & 83.00 & 82.00 & 80.00 & 83.43 \\
         FFBA (QG-Add) & 77.60 & 86.60 & 88.00 & 85.40 & 92.20 & 83.70 & 78.70 & 83.10 & 84.41 \\\hline
    \end{tabular}
    }
    \caption{Performance of different adapters on commonsense reasoning tasks.}
    \label{tab:cs}
\end{table*}
\paragraph{Fused forward-backward adapter}
In order to address this loss in performance or expressibility of the adapter module, we propose to fuse the backward adapter matrix $B$ into another linear projection matrix within the same block.
In a multi-head attention (MHA) block, the backward adapters for the query, key and value projections can be attached or fused into the output projection matrix.
As a result, all four projection layers within the MHA block have a forward adapter, while only the output projection layer has both forward as well as separate or shared backward adapters.
In a FFN block, all projection layers have a forward adapter while only the down-projection layer has both forward and backward adapters.
The resulting block-schematic of the fused forward-backward adapter (FFBA) or fully-fused low-rank adapter (FLoRA) within a transformer layer/block is show in Fig.~\ref{fig:ffba}.
The resulting projection layers with fused adapters are shown in Fig.~\ref{fig:ffba_blocks}, and are referred to as fused forward layer (FFL) and fused forward-backward layer (FFBL).

The motivation for fusing the forward and backward projections of the adapter into different projection layers come from parallel adapters where one single adapter per MHA or FFN block performs almost similar to LoRA with same parameter count~\cite{hu-etal-2023-llm}.
One thing that needs to be explored is the importance of the non-linearity operation used in parallel adapters at the bottleneck, and the placement of these forward and backward adapter projections.
The resulting computations for a layer with both forward and backward adapters attached is given by
\begin{align}
\small
\begin{bmatrix} Y \\ \Delta Y \end{bmatrix}
= \begin{bmatrix} W & B\\ A & C \end{bmatrix} \begin{bmatrix} X \\ \Delta X \end{bmatrix} 
= \begin{bmatrix} WX + B\Delta X\\ AX + C\Delta X \end{bmatrix}
\end{align}
The shrink operation reduces the output dimension from $(d+r)$ output to the original hidden embedding dimension $d$ of the base model.
If we remove the matrices $A$ and $C$ from the fused forward-backward layer it will become a purely fused-backward layer (FBL).
This also helps in removing the additional overhead of shrink (or repeat and add) operation.
The parameter count from $A$ can be added into $B$, however, the two operations are not equivalent and how they fare in terms of fine-tuning accuracies is a matter of investigation.
Also, it is to be noted that the FFBL or FBL needs an additional augmented input $\Delta X$, which comes from the augmented output $\Delta Z$ of the FFLs.
As shown in Fig.~\ref{fig:ffba}, the backward adapter fused into the output projection layer in MHA block is shared by the 3 forward adapters fused onto the QKV projections. Similarly, the backward adapter fused into the down-projection layer in FFN block is shared by the 2 forward adapters fused onto the up and gate projection layers. The augmented outputs of the FFLs are added together before feeding them as augmented input to the FBLs.

\section{Experiments and results}
Details of the experimental setup, datasets used, and the results are presented in this section.

\subsection{Datasets}
The performance of the proposed fused forward-backward adapters is evaluated on 3 different category of tasks, namely, commonsense reasoning, arithmetic reasoning and summary-dialogue generation.
For commonsense and arithmetic reasoning tasks, we use the Commonsense170K and Math10K training datasets used in ~\cite{hu-etal-2023-llm}.
For summary-dialogue tasks we use a combination of training sets from 4 different tasks, namely, CNN-DailyMail, Xsum~\cite{nallapati-etal-2016-abstractive}, DailyDialogue~\cite{li2017dailydialog}, and MultiWoz~\cite{budzianowski2018large}.
In order to speed up evaluations, all test sets in the commonsense reasoning tasks were limited to a maximum of top 1000 examples, to a maximum of top 500 examples for summary and dialogue tasks, and all examples of the math reasoning tasks.

\begin{table*}[h]
    \centering
    {
    \begin{tabular}{l|cccccc|c}
    \hline
     & \multicolumn{6}{c|}{Arithmetic Reasoning Tasks (Acc \%) } \\
         Adapter &  addsub & aqua & arith & gsm8k & singeq & svamp & Avg\\\hline
         \multicolumn{8}{c}{Llama3.2-1B-Inst} \\\hline
         Base & 68.10 & 22.83 & 62.17 & 45.49 & 80.91 & 53.20 & 55.45 \\
         FFT  & 85.32 & 22.83 & 96.17 & 48.52 & 90.94 & 66.70 & \bf 68.41 \\\hline
         LoRA & 82.78 & 24.80 & 94.33 & 45.87 & 89.76 & 64.50 & 67.00 \\
         FFA  & 81.77 & 20.08 & 85.17 & 36.24 & 84.84 & 58.60 & 61.11 \\
         FFBA (AorB) & 87.85 & 23.62 & 94.33 & 43.21 & 89.17 & 64.00 & \bf 67.15 \\         
         FFBA-Relu (AorB) & 87.85 & 25.98 & 96.00 & 36.62 & 92.52 & 55.90 & 65.81 \\
         FFBA (QG-Add) & 84.30 & 23.62 & 93.83 & 45.87 & 89.76 & 65.40 & \bf 67.13 \\
         FFBA (FPA) & 83.04 & 25.20 & 94.00 & 45.64 & 89.37 & 61.90 & 66.52 \\
         FFBA (FPA-Relu) & 85.82 & 26.77 & 93.17 & 43.75 & 90.35 & 57.90 & 66.29 \\\hline\hline
         \multicolumn{8}{c}{Llama3.2-3B-Inst} \\\hline
         Base & 91.14 & 24.80 & 93.17 & 76.88 & 93.90 & 87.60 & \bf 77.91 \\
         FFT  & 90.13 & 25.20 & 98.67 & 65.66 & 92.72 & 79.00 & 75.23 \\\hline
         LoRA & 93.16 & 27.17 & 96.67 & 67.10 & 95.87 & 82.50 & 77.07 \\
         FFA  & 87.59 & 21.26 & 96.00 & 66.87 & 92.13 & 80.30 & 74.02 \\         
         FFBA & 92.41 & 25.59 & 95.33 & 73.31 & 92.52 & 85.20 & \bf 77.39 \\
         FFBA-Relu & 93.11 & 25.61 & 95.10 & 72.31 & 91.52 & 85.00 & 77.10\\
         FFBA (QG-Add) & 90.13 & 33.86 & 97.33 & 69.45 & 94.88 & 80.00 & \bf 77.60 \\\hline
    \end{tabular}
    }
    \caption{Performance of different adapters on arithmetic reasoning tasks.}
    \label{tab:arith}
\end{table*}
\begin{table*}[h]
    \centering
    {
    \begin{tabular}{l|P{7ex}P{7ex}P{7ex}P{7ex}|P{7ex}}
    \hline
     & \multicolumn{4}{c|}{Summary/Dialogue Tasks ($R_{Lsum}$) } \\
         Adapter &  cnndm & dd & woz & xsum & Avg\\\hline
         \multicolumn{6}{c}{Llama3.2-1B-Inst} \\\hline
         Base    &  25.28 & 13.03 & 13.81 & 19.49 & 17.90 \\
         FFT    &  28.37 & 16.58 & 30.45 & 32.67 & {\bf 27.01}  \\\hline
         LoRA    & 25.06 & 15.06 & 28.27 & 27.27 & 23.91 \\
         FFA    & 25.05 & 14.93 & 24.53 & 24.38 & 22.22 \\
         FFBA (AorB)    & 26.62 & 16.70 & 30.32 & 30.82 & \bf 26.11 \\
         FFBA-Relu (AorB) & 27.09 & 15.57 & 26.66 & 28.65 & 24.49 \\
         FFBA (QG-Add)    & 26.24 & 19.67 & 29.65 & 29.38 & \bf 26.23 \\\hline\hline
         \multicolumn{6}{c}{Llama3.2-3B-Inst} \\\hline
         Base & 25.10 & 14.45 &  16.68 & 20.54 & 19.19 \\
         FFT & 29.23 & 25.85 &  29.66 & 37.63 & \bf 30.59 \\\hline
         Lora & 27.44 & 19.81 &  29.62 & 32.34 & 27.30 \\
         FFA    &  26.04 & 18.45 &  28.67 & 31.85 & 26.25 \\
         FFBA (AorB)    & 28.10 & 19.88 &  31.50 & 34.64 & \bf 28.53 \\
         FFBA-Relu (AorB) & 28.11 & 19.20 & 30.12 & 33.14 & 27.64 \\
         FFBA (QG-Add) & 28.71 & 20.39 & 30.87 & 35.72 & \bf 28.92 \\\hline
    \end{tabular}
    }
    \caption{Performance of different adapters on summary/dialogue tasks.}
    \label{tab:summdia}
\end{table*}
\begin{table*}[h]
    \centering
    {
    \begin{tabular}{l|cccc|cccc}
    \hline
     & \multicolumn{4}{c|}{Llama3.2-1B-Inst} & \multicolumn{4}{c}{Llama3.2-3B-Inst}  \\
         Adapter &  \#Param & TTFT & TPOT  & \% $\uparrow$ & \#Param & TTFT & TPOT & \% $\uparrow$ \\\hline
         Base	& - & 11.9 & 6.6 & - & - & 25.5 & 11.7 & - \\
         PEFT-LoRA & 22.5M & 17.2 & 11.7 & 77 & 48.6M & 33.4 & 19.4 & 65\\
         LoRA & 22.5M & 15.5 & 8.9 & \bf 35 & 48.6M & 31.9 & 15.2 & \bf 30 \\\hline 
         \multicolumn{9}{c}{Fused Adapters} \\\hline
         pf-LoRA & 22.5M & 15.3 & 8.3 & 26 & 48.6M & 32.3 & 13.8 & 18 \\
         FFA & 23M &  15.1 & 7.9 & 20 & 51.8M & 30.6 & 13.2 & 13 \\
         FFBA (A\&B) & 21M & 15.2 & 9.1 & 38 &55M & 31.9 & 15.0 & 28 \\
         FFBA (AorB) & 21M &  15.2 & 8.4 & 28 & 55M & 30.9 & 14.1 & 20 \\
         FFBA-Relu (AorB) & 21M & 15.3 & 8.9 & 35 & 55M & 31.0 & 14.9 & 27 \\
         FFBA (FPA, No-Add) & 21M & 14.3 & 7.7 & 16.6 & 55M & 29.6 & 12.7 & 9 \\
         FFBA (QG-Add) & 21M & 14.7 & 8.2 & \bf 24.2 & 55M & 30.5 & 13.5 & \bf 15 \\\hline
    \end{tabular}
    }
    \caption{Inference latencies (in ms) for different adapters (rank $r=32$) computed on an Nvidia H100 GPU using a common subset of 200 examples from the cnndm dataset. The percentage increase (\% $\uparrow$) in TPOT over the base model is also shown.}
    \label{tab:lat}
\end{table*}

\subsection{Experimental setup}
All experiments in this paper are conducted using the publicly available Llama3.2 family of LLM models~\cite{dubey2024llama, metaLlama32}.
The instruction fine-tuned variants of the models, namely, Llama3.2-1B-Inst and Llama3.2-3B-Inst are used.
The performance of the proposed FFBA adapters is compared against the popular LoRA adapters.
All adapters were fine-tuned for 10 epochs separately for each of the 3 category of tasks on a single node of 8 H100 GPUs with a global batch size of 1M tokens.
Seven different learning rates (LR) from $1e-6$ to $1e-3$ at equal intervals were explored for each of the adapters, with a constant LR scheduling.
The adapter checkpoints are saved at the end of each epoch and the best performing checkpoint on a validation set is used for final evaluation.
The validation set is created using 500 randomized examples held-out from the training set.
All fine-tuning experiments and evaluations were conducted using our custom implementation of adapters on top of HuggingFace transformers.

\subsection{Results}
The performance of the proposed LLM adapters on 3 important category of downstream tasks is presented in this section.
A comparison against the base model, full fine-tuning (FFT) and LoRA is provided.

\subsubsection{Commonsense reasoning}
The performance of different adapters for the Llama3.2-1B-Inst and Llama3.2-3B-Inst models on the popular commonsense reasoning tasks when fine-tuned using different adapters is given in Table~\ref{tab:cs}.
As can be seen from the results, full fine-tuning (FFT) of the models perform the best as compared to fine-tuning using adapters.
The proposed FFBA adapters performs almost similar or better than the popular LoRA adapters and come closest to full fine-tuning.

FFBA (AorB) denotes a variant of the FFBA architecture shown in Fig.~\ref{fig:ffba}, where the query, key and value projections have only the forward adapter ($A$) while the output and down projections have only the backward adapter ($B$).
Removing the forward adapter from a layer helps avoid the need for the repeat and add operation and thereby help reduce the latency marginally.
FFBA (QG-Add) denotes another variant where the adapter component $AX$ at the output of the fused-projection is added back into the base model component $WX$ only for the query and gate projection layers, thereby eliminating the repeat and add ops on key, value and down projection layers.
Removing all repeat and add operations would be a good idea from latency point of view, but that would result in no task specific additional learning to the attention and gating computations.
In view of this, we propose to retain only the repeat and add ops for the query and down projection layers.
It can be seen that this FFBA (QG-Add) variant provides the best results.

It can also be seen that the benefit of adding a non-linearity (FFBA-Relu) in between the forward and backward projections of the adapter is not very tangible.
On the contrary, it appears to under-perform the other adapters without any non-linearity which is a bit surprising and needs further investigation.
It can also be seen that the fused forward adapter (FFA) with only the forward down-projections ($AX$) performs worse, signifying the importance of the LRA constraint that helps in searching for a sparse solution in a larger dimension space.

\subsubsection{Math reasoning}
The performance of the adapters for the two Llama3.2 models on arithmetic reasoning tasks is given in Table~\ref{tab:arith}.
A similar trend follows, as was seen in the case of commonsense reasoning evaluations.
The proposed FFBA adapter performs similar or better than LoRA and closest to FFT, while the FFA under-performs all other adapters.
However, it was seen that the Llama3.2-3B base model performance for the math reasoning tasks gsm8k and svamp are already the best and none of the adapters including full-finetuning can improve upon the base model.
One possibility is that the instruction fine-tuned model is likely to be trained with several math reasoning instruction data, and the Math10K fine-tuning training set used in this paper is not adding any additional diversity or information.

The rows FPA and FPA-Relu denote a variant of FFBA adapters where the repeat and add op is removed completely from all layers.
The resulting adapter architecture (with or without the non-linearity) would be exactly same as the parallel-adapter outlined in ~\cite{hu-etal-2023-llm}, and can be termed as fused parallel adapter (FPA).
It can be seen that this FPA performs comparable to other adapters but marginally worse than other FFBA or LoRA adapters.
However, the FPA adapters provide one of the best latencies (as discussed in the next section) and can be a good alternative for latency critical applications.
Experiments with these FPA variants were carried out only for arithmetic reasoning tasks for illustration, and were not carried out for other tasks.

\subsubsection{Summary and dialogue generation}
Summary and dialogue generation is an important downstream application of LLMs, and the performance of various adapters on this category of tasks is shown in Table~\ref{tab:summdia}.
It can be seen from the results that the proposed FFBA adapters (AorB and QG-Add) perform significantly better than the conventional LoRA fine-tuning, and are the closest to full fine-tuning results.

\section{Analysis}
A comparison and discussion on the inference time latencies of different fused adapters as compared to the base model and the popular LoRA adapters is provided in this section.

\subsection{Latency of fused adapters}
The time-to-first-token (TTFT) and time-per-output-token (TPOT) latencies of different adapters as compared to the base model is shown in Table~\ref{tab:lat}.
The latencies are computed for Llama3.2 1B and 3B models over a common subset of 200 examples from the CNN-DailyMail evaluation set.
The latencies are computed by repeating each forward pass through the entire model two times for each decoded token and recording the time taken for the second pass.
The latencies are measured on a Nvidia H100 gpu with 80GB memory at full precision.
It can be seen from the table that inferencing using the off-the-shelf generalized PEFT-LoRA framework is optimized and shows relatively higher latency numbers.
In view of this, we evaluate all adapters using our custom implementation of all adapters without any unnecessary generalization overheads that may be required by PEFT-LoRA.

It can be seen from the table that fused adapters (FFBA-AorB and FFBA-QG-Add) reduce the TPOT overhead of LoRA (35\% wrt base model) by around 7-8\% for the 1B model and 7-11\% for 3B models.
Equivalently, they reduce the LoRA TPOT overhead by 21-30\% (1B) and 31-48\% (3B) relatively.
FFBA (AorB) and FFBA (A\&B) in the table refers to the special case of FFBA where only the backward adapters or both forward and backward adapters are used alongside output- and down- projections in MHA and FFN blocks, respectively.
As can be seen from the latency figures using both forward and backward adapters increases the overhead due to the need for an extra repeat and add (shrink) operation.
Also, based on initial experiments conducted using FFBA-FB which did not show any better performance compared to FFBA-B, all experiments reported in the previous section are for the FFBA-B variant.
Also, using a non-linearity (FFBA-Relu) when using FFBA-B style adapters also increases the latency considerably.


\section{Conclusions}
In this paper, we proposed a family of fused adapters, FLoRA, which allows us to optimize the way adapter compute is handled at inference time.
The main idea is to fuse the smaller adapter projections into the adjacent large projection layers whenever possible.
This reduces the number of independent or sequential operations and there by reducing the GPU inefficiency in handling smaller matrix multiplications.
This results in reducing the LoRA TPOT latency overhead by around 21-30\% relatively for 1B models, and 31-48\% for 3B models.
In terms of performance, the FFB adapters are comparable or marginally better than LoRA for commonsense and math reasoning tasks.
However, they show significant improvement over LoRA for summary and dialogue tasks.

\section{Limitations}
We recognize the following limitations of our work.
The comparison of our proposed fused adapters is limited only to the popularly used LoRA adapters, and 3 different category of tasks.
A more exhaustive comparison against other adapters such as prefix adapters, and more downstream tasks can be part of future scope.
The experiments and down-stream applications considered in this paper are restricted to one language (English), one modality (text) and can be extended to other languages and modalities.
The experiments are limited to small or moderately sized LLMs that could be candidates for ondevice deployment.
Experiments with huge cloud based LLMs is not within the scope of this study.
The latency measurements in this paper are made on a H100 GPU using full precision.
Measurements on different GPUs and NPUs (for ondevice deployment of LLMs) at different precisions, as well as studying the effect of different quantization methods on these adapter accuracies are part of our future scope.
The role and significance of different types of non-linearity within the adapter architecture is not fully explored.
In order to speed up evaluations, all test sets in the commonsense reasoning tasks were limited to a maximum of top 1000 examples, to a maximum of top 500 examples for summary and dialogue tasks, and all examples of the math reasoning tasks. 
We believe this is a fair number to compare different methods.
Results on the full evaluation sets will be provided in the future versions, either as part of the main paper or as appendix.

\bibliography{custom, mybib}


\end{document}